\documentclass[conference]{IEEEtran}
\usepackage{times}

\usepackage[numbers]{natbib}
\usepackage{multicol}
\usepackage[bookmarks=true]{hyperref}

\usepackage{amsmath}
\usepackage{amsfonts}
\usepackage{amsthm}
\usepackage{amssymb}
\usepackage{mathrsfs}  
\usepackage{bm}
\usepackage[ruled,vlined]{algorithm2e}
\usepackage{dsfont} 
\usepackage{accents} 
\usepackage{tikz-cd} 
\usepackage{adjustbox} 
\usepackage{mathtools} 

\let\labelindent\relax
\usepackage{enumitem}

\usepackage{listings}
\usepackage{xcolor}
\usepackage{caption}
\usepackage{float}

\definecolor{codebg}{rgb}{0.95,0.95,0.95}
\definecolor{keywordcolor}{rgb}{0.0,0.0,0.6}   
\definecolor{typecolor}{rgb}{0.55,0.0,0.55}    
\definecolor{stringcolor}{rgb}{0.6,0.1,0.1}    
\definecolor{commentcolor}{rgb}{0.0,0.5,0.0}   
\definecolor{funcnamecolor}{rgb}{0.0,0.3,0.7}  

\usepackage{siunitx}

\usepackage{graphicx}

\usepackage{balance}

\newcommand{\comment}[1]{} 
 
\newcommand\parens[1]{\left(#1\right)} 

\usepackage{pifont}
\newcommand{\cmark}{\ding{51}}%
\newcommand{\xmark}{\ding{55}}%

\usepackage{lipsum}
\usepackage{booktabs}

\begin{document}

\title{Judo: A User-Friendly Open-Source Package for \\Sampling-Based Model Predictive Control}

\author{
    Albert H. Li$^{*, \dagger}$, Brandon Hung$^{\dagger}$, Aaron D. Ames$^{*}$ \\
    Jiuguang Wang$^{\dagger}$, Simon Le Cleac'h$^{\dagger,\ddagger}$, and Preston Culbertson$^{\dagger,\ddagger}$ \\
    $^*$ Caltech, $^\dagger$ RAI Institute, $\ddagger$ Equal Advising \\
    \{alberthli, ames\}@caltech.edu, \{bhung, jw, slecleach, pculbertson\}@rai-inst.com
}

\maketitle

\begin{abstract}
Recent advancements in parallel simulation and successful robotic applications are spurring a resurgence in sampling-based model predictive control. To build on this progress, however, the robotics community needs common tooling for prototyping, evaluating, and deploying sampling-based controllers. We introduce \verb|judo|, a software package designed to address this need. To facilitate rapid prototyping and evaluation, \verb|judo| provides robust implementations of common sampling-based MPC algorithms and standardized benchmark tasks. It further emphasizes usability with simple but extensible interfaces for controller and task definitions, asynchronous execution for straightforward simulation-to-hardware transfer, and a highly customizable interactive GUI for tuning controllers interactively. While written in Python, the software leverages MuJoCo as its physics backend to achieve real-time performance, which we validate across both consumer and server-grade hardware. Code at \href{https://github.com/bdaiinstitute/judo}{https://github.com/bdaiinstitute/judo}.

\end{abstract}

\IEEEpeerreviewmaketitle

\section{Introduction}\label{sec:intro}


Recent advances in parallel model-based simulation tools have shown the effectiveness of \textit{sampling-based} algorithms like predictive sampling, the cross-entropy method (CEM), model predictive path integral control (MPPI), and more for generating rich, dynamic plans for a wide variety of tasks (including contact-rich ones) in real time with limited resources \cite{howell2022_mjpc}. Moreover, recent results demonstrate that these strategies can effectively solve real-world tasks, including quadrupedal and bipedal locomotion \citep{alvarezpadilla2024realtimewholebodycontrollegged, xue2024fullordersamplingbasedmpctorquelevel, zhang2025wholebodymodelpredictivecontrollegged} as well as dexterous, in-hand object reorientation \cite{li2025_drop}. However, if the robotics community wishes to build upon these successes, simple and effective open-source software is critical for further investigating and deploying these methods. 

To that end, this work presents \verb|judo|, an open-source library for developing, tuning, and deploying sampling-based MPC algorithms. The aim of \verb|judo| is to provide a user-friendly implementation that is also performant, and enables simple transfer of controllers from simulation to hardware.

\section{Design Principles}

\begin{figure}[t]
    \centering
    \includegraphics[width=\linewidth]{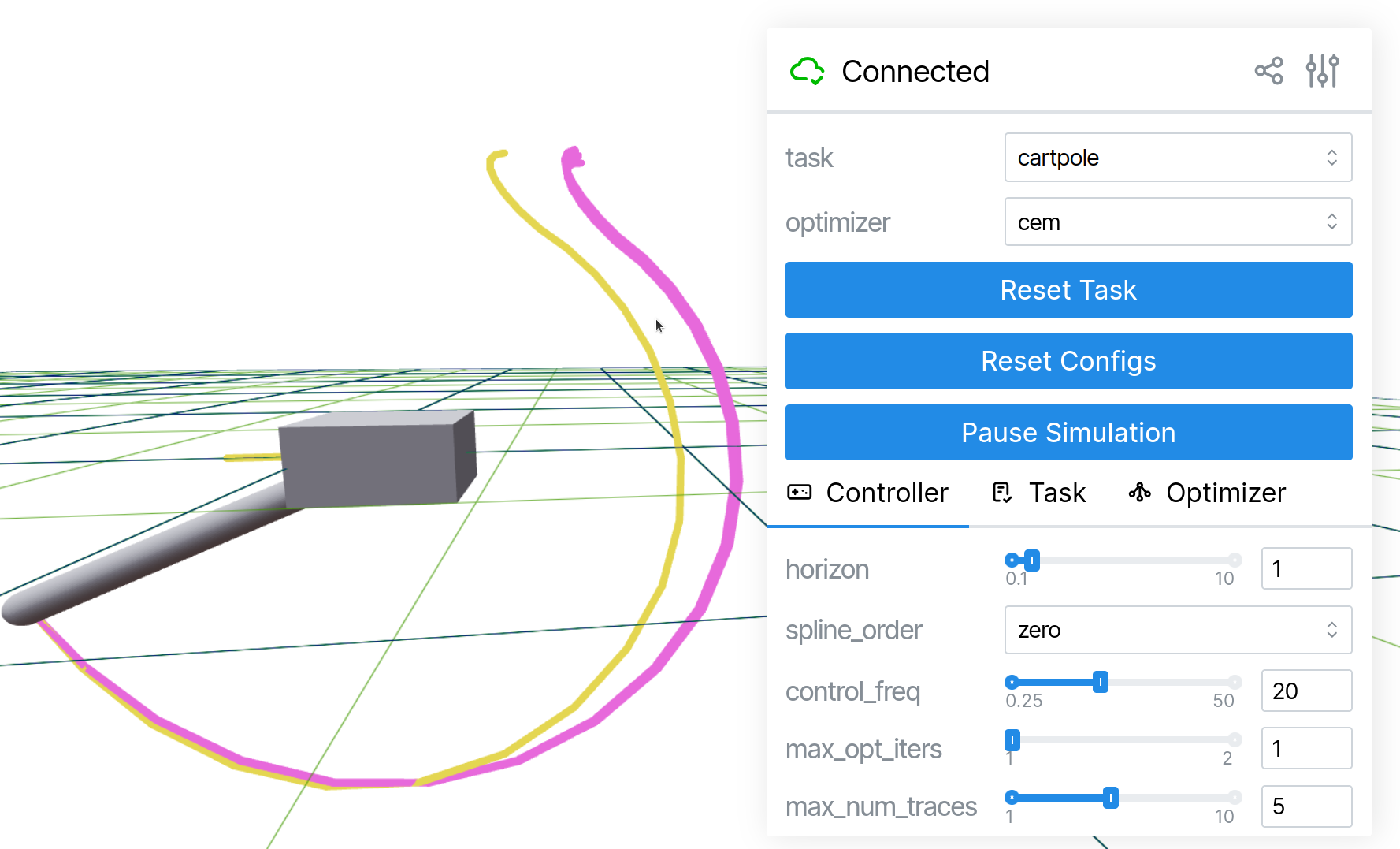}
    \caption{The \texttt{judo} interface. The GUI is interactive, allowing users to tune parameters in real time. Dropdown menus allow switching between different tasks and controllers with ease.}
    \label{fig:judo}
\end{figure}

\begin{figure*}[t]
\centering
\begin{minipage}{0.95\linewidth}
\begin{lstlisting}[caption={Minimal example implementations of a Task and an Optimizer using the \texttt{judo} API.}, label={lst:examples}]
# Example Task
@dataclass
class CartpoleConfig(TaskConfig):
    w_vert: float = 10.0
    w_ctr: float = 10.0
    w_vel: float = 0.1
    w_ctrl: float = 0.1

class Cartpole(Task[CartpoleConfig]):
    def __init__(self, model_path=XML_PATH):
        super().__init__(model_path, sim_model_path=XML_PATH)
        self.reset()

    def reward(self, states, sensors, controls, config, system_metadata=None):
        x, y, vel = states[..., 0], states[..., 1], states[..., 2:]
        vertical_rew = -config.w_vert * smooth_l1_norm(np.cos(y) - 1, 0.01).sum(-1)
        centered_rew = -config.w_ctr * smooth_l1_norm(x, 0.1).sum(-1)
        velocity_rew = -config.w_vel * quadratic_norm(vel).sum(-1)
        control_rew = -config.w_ctrl * quadratic_norm(controls).sum(-1)
        return vertical_rew + centered_rew + velocity_rew + control_rew

    def reset(self) -> None:
        self.data.qpos = np.array([1.0, np.pi]) + np.random.randn(2)
        self.data.qvel = 1e-1 * np.random.randn(2)
        mujoco.mj_forward(self.model, self.data)

# Example Optimizer
@dataclass
class PredictiveSamplingConfig(OptimizerConfig):
    sigma: float = 0.05

class PredictiveSampling(Optimizer[PredictiveSamplingConfig]):
    def __init__(self, config: PredictiveSamplingConfig, nu: int) -> None:
        super().__init__(config, nu)

    def sample_control_knots(self, nominal_knots):
        nn = self.num_nodes
        nr = self.num_rollouts
        sigma = self.config.sigma
        noised_knots = nominal_knots + sigma * np.random.randn(nr - 1, nn, self.nu)
        return np.concatenate([nominal_knots, noised_knots])

    def update_nominal_knots(self, sampled_knots, rewards):
        i_best = rewards.argmax()
        return sampled_knots[i_best]
\end{lstlisting}
\end{minipage}
\end{figure*}
The design of \verb|judo| was driven by three core principles:
\begin{itemize}
\item \textbf{Maximize Research Velocity.} Echoing MJPC \cite{howell2022_mjpc}, \verb|judo|'s main goal is to accelerate research in sampling-based MPC. To achieve this, \verb|judo| is implemented in Python, offering an effective balance of rapid prototyping and performance suitable for rapid development. Its simple yet extensible interfaces are designed to facilitate extension of (and contribution to) the core codebase. Moreover, an integrated GUI provides real-time visualization and tuning of task and algorithm hyperparameters, significantly shortening the development loop.

\item \textbf{Accurate and Fast Simulation.} While user-friendliness is a priority, \verb|judo| also requires a high-performance simulation engine for effective controller development and real-time applications. It utilizes MuJoCo as its primary modeling backend, enabling detailed simulation of complex scenes, geometries, and dynamics as defined via MuJoCo XML, and leveraging MuJoCo’s highly-optimized, multi-threaded C-based forward dynamics. Crucially, \verb|judo| is designed with a modular simulation backend interface; this allows researchers to integrate alternative physics engines or their own custom dynamics models if their work requires it.

\item \textbf{Streamline Sim-to-Real Transfer.} Recognizing that discrepancies between simulation and hardware are often major obstacles to hardware deployment, \verb|judo| is engineered to minimize sim-to-real gaps and simplify the transition to hardware. The core application architecture separates control logic from physics simulation, running them in asynchronous threads. This design allows for straightforward hardware deployment: researchers can develop a specific hardware interface and substitute it for the simulation component with minimal changes to their controller code. This maximizes code reuse and provides an accurate preview of the control stack's performance characteristics prior to deployment on physical hardware.
\end{itemize}

Numerous other open-source libraries exist for developing and deploying sampling-based MPC algorithms (see Table \ref{tab:sampling_libs}).

\begin{table}[h]
\caption{Comparison of sampling-based MPC libraries}
\label{tab:sampling_libs}
\centering
\setlength{\tabcolsep}{4pt}

\begin{tabular}{|c||c|c|c|c|c|}
\hline
Library & Lang & GUI & Sim & Multi-Planner & Async \\
\hline
\hline
MJPC \cite{howell2022_mjpc} & C++ & \cmark & MuJoCo & \cmark & \cmark \\
hydrax \cite{kurtz2024hydrax} & Python & \xmark & MJX & \cmark & \cmark \\
MPPI-Generic \cite{vlahov2024mppigenericcudalibrarystochastic} & C++ & \xmark & Manual & \xmark & \xmark \\
mppi-isaac \cite{pezzato2023_isaac_mppi} & Python & \cmark & Isaac Gym & \xmark & \xmark \\ 
pytorch\_mppi \cite{pytorch_mppi} & Python & \xmark & Manual & \xmark & \xmark \\
\hline
Judo (ours) & Python & \cmark & MuJoCo & \cmark & \cmark \\
\hline
\end{tabular}

\end{table}

Besides MJPC and \texttt{hydrax}, the majority of these libraries have fairly limited scope, like only supporting one planning algorithm (e.g., MPPI) or only exposing a limited interface for dynamical systems, requiring the user to manually write the dynamics. Thus, most existing software tooling for sampling-based MPC is less extensible towards wider algorithms or systems compared to \verb|judo|. \texttt{hydrax}, an MJPC-inspired toolkit in Python, uses MJX \cite{mjx}, a \verb|jax|-based re-implementation of MuJoCo as its simulation backend, and does not provide GUI functionality for controller or task tuning.  While inspired by MJPC, \verb|judo| is significantly easier to install or use as a dependency downstream.
The remainder of this paper will provide detail on some of \verb|judo|'s design decisions. The open-source code is available at \href{https://github.com/bdaiinstitute/judo}{github.com/bdaiinstitute/judo}. Of course, nothing replaces simply ``playing with'' the package:
\begin{verbatim}
pip install judo-rai  # one-line install
judo  # runs the browser-based GUI
\end{verbatim}

\section{The \texttt{judo} Package}\label{sec:judo}

\subsection{The Controller Interface}\label{sec:interface}
We assume sampling-based MPC controllers to have the structure shown in Alg. \ref{alg:sampling_planner}. The nominal control signal that is executed on the system is parameterized by the knots $\theta$ of some control spline, which we interpolate at time $t$.

\SetKwComment{Comment}{// }{}
\begin{algorithm}[h]

\caption{Sampling-based MPC \cite{li2025_drop}}
\label{alg:sampling_planner}
\KwIn{$\theta$, $N$, planner-specific parameters.}

\While{planning}{
    $x_0 \leftarrow \hat{x}(t)$ \Comment*[r]{estimate curr state}
    \For{$i=1$ to $N$ \Comment*[r]{multi-threaded}}{
        $U^{(i)} \sim \pi_{\theta}(U)$ \Comment*[r]{sample controls}
        $J^{(i)} \gets J\left(U^{(i)}; x_0\right)$ \Comment*[r]{eval rollout}
    }
    $\theta \leftarrow \mathtt{update\_params}\parens{U^{(1:N)}, J^{(1:N)}}$;
    
    $u(t) \gets \mathtt{get\_action}(\theta, t)$ \Comment*[r]{asynchronous}
}
\end{algorithm}

At the core of the \verb|judo| API are \textit{tasks} and \textit{optimizers}, which are sub-components of the above controller. The task specifies the reward, which allows us to evaluate the rollouts. The optimizer specifies the procedure for sampling new control spline knots as well as how we should update the nominal policy given the samples and their corresponding rewards. Minimal examples of task and optimizer implementations are shown in Listing \ref{lst:examples}.

We provide a number of starter tasks, ranging from low-dimensional examples like the cartpole system to more complex examples, including pick and place and cube rotation. We also supply some initial implementations of canonical sampling-based control algorithms, including predictive sampling, CEM, and MPPI. Lastly, we provide an extensible abstract interface for allowing alternative rollout backends in upcoming releases, such as \verb|mujoco_warp| \cite{mujoco_warp} or custom dynamics.

\subsection{The GUI}\label{sec:viz}

\begin{figure}[t]
    \centering
    \begin{minipage}{0.95\linewidth}
    \begin{lstlisting}
@slider("num2", 0.0, 10.0, 0.1)
@dataclass
class DummyOptimizerConfig(OptimizerConfig):
    num1: int = 42  # default slider
    num2: float = 3.14  # custom slider
    num3: float = 2.71  # default slider
    checkbox: bool = True
    options: Literal["opt1", "opt2"] = "opt1"
    arr: np.ndarray = np_1d_field(
        np.array([1.0, 2.0]),
        names=["field1", "field2"],
        mins=[0.0, 1.0],
        maxs=[10.0, 20.0],
        steps=[0.1, 0.2],
    )
    \end{lstlisting}
    \end{minipage}
    \includegraphics[width=0.75\linewidth, trim=0 0 0 8.25cm, clip]{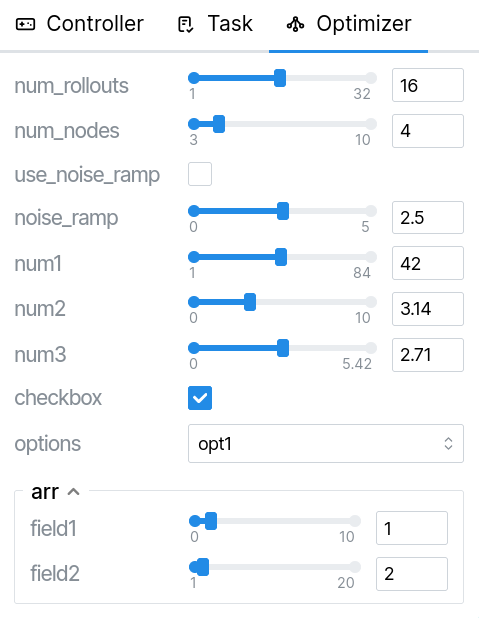}
    \caption{A config and the corresponding automatically-generated GUI elements. Note the \texttt{slider} decorator, which allows fine-grained control over slider limits and step size.}
    \label{fig:gui_elements}
\end{figure}

The \verb|judo| GUI is built on \verb|viser| \cite{viser2023}, a highly-customizable browser-based visualizer. In \verb|judo|, the fields of registered task or optimizer config objects are automatically parsed into the appropriate GUI elements. For example, consider the dummy optimizer config in Fig. \ref{fig:gui_elements}. Integer and float fields are parsed as sliders, booleans as checkboxes, literals as dropdown menus, and numpy arrays as folders with subsliders. Moreover, \verb|judo| provides a flexible decorator-based interface for more finely controlling the range and step size of sliders.

To visualize the optimization iterates, \verb|judo| allows the user to specify traces in the model XML descriptions. To do so, \verb|judo| simply looks for sensors of type \verb|framepos| with the substring ``trace'' in their name. Then, the nominal and a few of the highest-performing traces (the number of which can be adjusted) are shown in the visualizer over the rollout horizon.

Lastly, \verb|viser| allows active browser sessions to be shared with anyone else. This means that collaborators can view and even interact with a live session from anywhere in the world along with the user.

\subsection{Asynchronous Operation}\label{sec:async}
The \verb|judo| stack is modular, consisting of a ``system'' node (e.g., the simulator or whatever nodes are required to run a hardware stack), a visualizer node, and a controller node. These nodes all asynchronously communicate with each other, where the interprocess communication is handled by \verb|dora| \cite{dora2025}, a fast robotics-first middleware built on \verb|rust| with a full-featured Python interface.

We find that \verb|dora| is substantially easier to install, set up, and use than other frameworks like ROS2, and the entire communication graph is natively specified in a standardized YAML format. Moreover, \verb|dora| allows zero-copy reads of array data between nodes on the same device, which includes any array built on the \verb|dlpack| specification as well as \verb|pytorch| arrays on both CPU and GPU. If the simulation node is swapped for hardware nodes, then the same visualizer and controller nodes can be used, which facilitates low-friction transfer between simulation and hardware.

\subsection{Registration and Configuration Management}\label{sec:config}

\begin{figure}[t]
\centering
\begin{minipage}{0.95\linewidth}
\begin{lstlisting}[caption={Example of custom registration and task-specific overrides using either a python script or hydra YAML.}, label={lst:registration_and_overrides}]
# Example: programmatic registration/overrides
from judo.cli import app
from judo.config import set_config_overrides
from judo.optimizers import register_optimizer
from judo.tasks import register_task

if __name__ == "__main__":
    register_optimizer("my_opt", MyOpt, MyOptCfg)
    register_task("my_task", MyTask, MyTaskCfg)
    set_config_overrides(
        "my_task",
        ControllerConfig,
        {"horizon": 1.0}
    )
    set_config_overrides(
        "cylinder_push",
        MyOptCfg,
        {"param": 42},
    )
    app()

# Example: YAML-based registration/overrides
defaults:
  - judo  # must use this default!
task: "my_task"  # init task in gui
custom_tasks:
  my_task:
    task: module.path.to.MyTask
    config: module.path.to.MyTaskCfg
custom_optimizers:
  my_opt:
    optimizer: module.path.to.MyOpt
    config: module.path.to.MyOptCfg
controller_config_overrides:
  my_task:
    horizon: 1.0
optimizer_config_overrides:
  cylinder_push:
    my_opt:
      param: 42
\end{lstlisting}
\end{minipage}
\end{figure}

There are many parameters to configure in the full \verb|judo| stack, including the parameters of each node, as well as task-specific overrides to controller/optimizer configuration defaults. For example, in Task A, we may want a planning horizon of 1 second, whereas for Task B, we may want to use 0.5 seconds. Both of these may also differ from some default value for the controller of interest, say, 0.75 seconds. Most importantly, when a user writes a custom controller or task, it must be \textit{registered} into \verb|judo| so that it appears in the interactive dropdown menus in the GUI.

We designed \verb|judo| such that registration and configuration management could occur in one of two ways for most cases: (1) programmatically, by manually defining and registering parameters and/or overrides in Python, and (2) with YAML syntax via \verb|hydra| \cite{Yadan2019Hydra}, where only a single file need be written and passed to the main \verb|judo| executable.

For example, in Listing \ref{lst:registration_and_overrides}, we can register custom tasks/optimizers and specify task-specific optimizer overrides in a python script, or we can identically specify the same changes in a \verb|hydra| YAML file located at \verb|/path/to/folder/example.yaml|, which we can pass to \verb|judo| with the command:
\begin{verbatim}
judo -cp /path/to/folder -cn example
\end{verbatim}

\section{Evaluation}\label{sec:evaluation}
In the sampling-based MPC loop, the computational bottleneck is the nominal control knot update (and in particular, the multi-threaded rollouts). To evaluate whether \verb|judo| is viable for real-time control, we provide speed benchmarks for the control update on a consumer-grade ThinkPad X1 Carbon Gen 11 laptop with a 12-thread 13th generation i7 CPU, as well as a server-grade workstation with an 128-thread Ryzen Threadripper Pro 5995wx CPU.

We test three tasks: the canonical \verb|cartpole| task; the \verb|cylinder_push| task, where an actuated cylinder must push a target cylinder to a goal location; and the \verb|leap_cube| task, where a LEAP hand must rotate a cube to reach as many consecutive goal orientations in a row as possible. For each of these three tasks, we test the speed of three optimizers: predictive sampling, CEM, and MPPI. All benchmark speeds are reported using 10 planning threads, and the threadripper is additionally profiled at 120 threads to show that the speed can be maintained at a high thread count.

\begin{table}[h]
\centering
\caption{Avg. control update timing (ms).}
\begin{tabular}{lccc}
\toprule
\multicolumn{4}{c}{Task: \texttt{cartpole}} \\
\midrule
& CEM & MPPI & PS \\
\midrule
ThinkPad (10 threads) & $4.4 \pm 0.7$ & $4.4 \pm 0.5$ & $4.3 \pm 0.6$ \\
Threadripper (10 threads) & $1.3 \pm 0.2$ & $1.5 \pm 0.2$ & $1.3 \pm 0.2$ \\
Threadripper (120 threads) & $1.4 \pm 0.2$ & $1.3 \pm 0.2$ & $1.3 \pm 0.2$ \\
\midrule
\multicolumn{4}{c}{Task: \texttt{cylinder\_push}} \\
\midrule
& CEM & MPPI & PS \\
\midrule
ThinkPad (10 threads) & $5.3 \pm 0.8$ & $7.1 \pm 1.3$ & $6.7 \pm 1.5$ \\
Threadripper (10 threads) & $2.1 \pm 0.5$ & $2.5 \pm 1.1$ & $2.0 \pm 0.3$ \\
Threadripper (120 threads) & $2.0 \pm 0.3$ & $2.3 \pm 0.2$ & $2.1 \pm 0.3$ \\
\midrule
\multicolumn{4}{c}{Task: \texttt{leap\_cube}} \\
\midrule
& CEM & MPPI & PS \\
\midrule
ThinkPad (10 threads) & $48.8 \pm 4.0$ & $51.7 \pm 5.8$ & $51.2 \pm 5.2$ \\
Threadripper (10 threads) & $19.8 \pm 2.9$ & $17.7 \pm 2.1$ & $16.8 \pm 1.8$ \\
Threadripper (120 threads) & $15.4 \pm 1.5$ & $15.3 \pm 1.8$ & $14.5 \pm 1.4$ \\
\bottomrule
\end{tabular}
\end{table}

We observe that the Threadripper computes rollouts about 2.5 times faster than the i7, though both are quick enough to run in real time. The disparity between consumer-grade and server-grade hardware is a limitation of sampling-based MPC, as noted in prior work \cite{li2025_drop}. However, as experimental GPU-based parallel simulators like \verb|mujoco_warp| stabilize, we expect that this disparity will shrink.

\section{Conclusion}\label{sec:conclusion}
We hope that \verb|judo| can facilitate the prototyping and development of sampling-based MPC algorithms in the wider robotics community. By focusing on a feature-rich and user-friendly interface, \verb|judo| provides a hackable, extensible, and expressive framework for the development of new algorithms and investigation of complex tasks. In the future, we have an exciting roadmap of upcoming features, including integration with learned controllers, examples of hardware usage, and adding many more tasks and optimizers to the core library.


\clearpage
\newpage
\balance

\bibliographystyle{plainnat}
\bibliography{references}

\end{document}